# Investigation of Compressor Cascade Flow Using Physics-Informed Neural Networks with Adaptive Learning Strategy

Zhihui Li,[*]  Francesco Montomoli[†] and Sanjiv Sharma[‡]
*UQ Laboratory, Department of Aeronautics, Faculty of Engineering, Imperial College London, London, SW7 2AZ, UK*

**In this study, we utilize the emerging Physics Informed Neural Networks (PINNs) approach for the first time to predict the flow field of a compressor cascade. Different from conventional training methods, a new adaptive learning strategy that mitigates gradient imbalance through incorporating adaptive weights in conjunction with dynamically adjusting learning rate is used during the training process to improve the convergence of PINNs. The performance of PINNs is assessed here by solving both the forward and inverse problems. In the forward problem, by encapsulating the physical relations among relevant variables, PINNs demonstrate their effectiveness in accurately forecasting the compressor's flow field. PINNs also show obvious advantages over the traditional CFD approaches, particularly in scenarios lacking complete boundary conditions, as is often the case in inverse engineering problems. PINNs successfully reconstruct the flow field of the compressor cascade solely based on partial velocity vectors and near-wall pressure information. Furthermore, PINNs show robust performance in the environment of various levels of aleatory uncertainties stemming from labeled data. This research provides evidence that PINNs can offer turbomachinery designers an additional and promising option alongside the current dominant CFD methods.**

## Nomenclature

AD  =  Automatic Differentiation

CFD  =  Computational Fluid Dynamics

---

[*] Research Fellow, Department of Aeronautics, Faculty of Engineering, Imperial College London, Email: zhihui.li@imperial.ac.uk
[†] Professor, Department of Aeronautics, Faculty of Engineering, Imperial College London, Email: f.montomoli@imperial.ac.uk
[‡] Professor, Department of Aeronautics, Faculty of Engineering, Imperial College London, Email: sanjiv.sharma2021@gmail.com

| | | |
|---|---|---|
| CNNs | = | Convolutional Neural Networks |
| DNNs | = | Deep Neural Networks |
| DNS | = | Direct Numerical Simulation |
| GPU | = | Graphics Processing Unit |
| HPC | = | High Performance Computing |
| LES | = | Large Eddy Simulation |
| NN | = | Neural Network |
| N-S | = | Navier-Stokes |
| PDEs | = | Partial Differential Equations |
| PINNs | = | Physics-informed Neural Networks |
| PIV | = | Particle Imaging Velocimetry |
| RANS | = | Reynolds-averaged Navier-Stokes |
| SA | = | Spalart Allmaras |
| $MSE$ | = | Mean Square Error |
| $T$ | = | Epoch steps |
| $\mathcal{L}$ | = | Loss Function |
| $\mathcal{L}_b$ | = | Training Loss on Computation Boundaries |
| $\mathcal{L}_e$ | = | Residual Loss of Governing Equations |
| $\mathcal{L}_f$ | = | Training Loss on Training Points inside Flowfield |
| $N_e$ | = | Number of Sample Points |
| $L^2$ | = | Quadratic Norm |
| $R^2$ | = | R-squared Value |
| $C_p$ | = | Static Pressure Coefficients |
| $\mathcal{O}$ | = | Order of Magnitude |
| $n$ | = | Number of Decay Cycles |
| $p$ | = | Static Pressure |
| $u$ | = | Axial Component of Velocity |
| $v$ | = | Vertical Component of Velocity |
| $x$ | = | Axial Coordinates |
| $y$ | = | Vertical Coordinates |
| $lr$ | = | Learning Rate |
| $\omega_i$ | = | Weights of Loss Terms |
| $\Omega$ | = | Space Coordinates Set |
| $\boldsymbol{f}$ | = | Residual Function of N-S Equations |
| $\mathbf{u}$ | = | Velocity Vector |
| $\boldsymbol{\lambda}$ | = | Other Parameters |
| $\mathbf{x}$ | = | 2D Coordinate Space |
| $\boldsymbol{I}$ | = | Initial Condition |
| $\boldsymbol{B}$ | = | Boundary Conditions |
| $\boldsymbol{\theta}$ | = | Trainable Parameters |
| $v$ | = | Kinematic Viscosity of Air |
| $v_{tur}$ | = | Turbulent Viscosity |
| $\nabla$ | = | Gradient Operator |
| $\nabla^2$ | = | Laplace Operator |
| $\partial$ | = | Partial Differential Operator |
| $\mathbb{R}$ | = | Real Number Field |
| $y^+$ | = | Normalized Distribution between First-Layer Grid and Wall |
| 2D | = | Two-Dimensional |

*Subscripts*

| | | |
|---|---|---|
| $NN$ | = | Neural Network |
| $b$ | = | Boundary Points |
| $e$ | = | Residual Points |
| $f$ | = | Labeled data |
| $i$ | = | *i-th* |
| $in$ | = | Inlet Surface |
| *lower* | = | Lower Periodic Surface |
| *max* | = | Maximum |
| *min* | = | Minimum |
| *nw* | = | Near Wall |
| *out* | = | Outlet Surface |
| *per* | = | Periodic Surfaces |
| *s* | = | Static |
| *t* | = | Total |
| *tur* | = | Turbulent |
| *upper* | = | Upper Periodic Surface |
| *wall* | = | Wall Surface |

*Subscripts*

| | | |
|---|---|---|
| *i* | = | *i-th* |
| − | = | Mean Parameters |

## I. Introduction

Physics-informed neural networks (PINNs) have recently emerged as a promising approach for solving problems in engineering, communication, and biology [1-3]. The fundamental concept behind PINNs is to incorporate knowledge of the governing physical laws into the learning process of neural networks (NNs). By leveraging the universal approximation theory of neural networks [4], incorporating prior physical information into the neural network can effectively constrain the solution space and demonstrate strong generalization capabilities across different inputs.

The Navier-Stokes (N-S) equations capture the physical laws governing various engineering flow systems. In traditional scenarios [5-7], solutions to the N-S equations are approximated by incorporating appropriate initial and boundary conditions along with various space-time discretization methods. Deep neural networks (DNNs) provide an alternative approach for approximating solutions to the N-S equations. However, these networks rely heavily on the size of the training dataset due to the lack of prior knowledge about the governing equations and there are no constraints that guarantee a physical correlation between the quantities of interest. PINNs, on the other hand, are introduced as a new class of numerical solvers for diverse flow problems, utilizing the available training data (even if sparse) and the imposed N-S equations as guidance.

Raissi et al. [8] were the first to propose the use of PINNs for solving nonlinear partial differential equations governing incompressible flow around a circular cylinder. The PINNs were able to produce a qualitatively accurate pressure field using only velocity information from the training dataset. Raissi et al. [9] then developed a PINN framework to extract quantitative velocity and pressure information for both external and internal flow scenarios. Jin et al. [10] applied PINNs with velocity-pressure and vorticity-velocity formulations to approximate laminar and turbulent incompressible flows, and the results exhibited reasonable agreement with analytical values. Eivazi et al. [11] utilized PINNs to solve Reynolds-averaged Navier-Stokes (RANS) equations for laminar and turbulent flows, demonstrating good prediction accuracy for the NACA4412 flow field, including Reynolds-stress components.

Ryck et al. [12] conducted a theoretical investigation into the behavior of total and training losses of PINNs when approximating incompressible Navier-Stokes equations. They proved that the total L2 loss of PINNs was bounded by the residuals of the partial differential equations (PDEs) of Navier-Stokes equations, and reducing the total prediction error of PINNs required sufficient sampling of residual points. Arthurs and King [13] approximated solutions of the Navier-Stokes equations across the design parameter space using a combination of PINNs and active learning approaches. The use of PINNs significantly reduced the required data storage space compared to conventional CFD data files. Xiang et al. [14] employed Gaussian probabilistic models to determine the weights of loss terms in PINNs and concluded that self-adaptive weighted PINNs exhibited better performance in modeling incompressible flows compared to traditional PINNs with fixed weights of loss terms. Du et al. [15] investigated the spatial-temporal interpolation inside minimal turbulent channel flow using PINNs and compared the predictive performance of PINNs with the adjoint-variational data assimilation method. Cai et al. [16] provided a comprehensive review of various applications of PINNs for inverse problems related to fluid dynamic systems, including incompressible cylinder flow, compressible bow shock waves, and biomedical flow.

Currently, the design and optimization of gas turbines heavily rely on various Computational Fluid Dynamics (CFD) methods. However, the use of high-fidelity CFD approaches like Large Eddy Simulation (LES) [17] and Direct Numerical Simulation (DNS) [18] is limited in engineering applications due to their high computational costs. As an alternative to CFD methods, PINNs offer unique advantages by bypassing the laborious processes of geometric reconstruction, mesh generation, and numerical discretization. Consequently, the utilization of PINNs for solving gas turbine flows has become a burgeoning topic in the field. In a recent study, Post et al. [19] employed PINNs to approximate the LS89 turbine nozzle by solving the steady 2D Euler equations. The results showcased the potential of PINNs in uncovering hidden fluid mechanics, particularly in the context of inverse problems. Another application of PINNs in gas turbines was demonstrated by Lyathakula et al. [20], who utilized them to resolve the Elasto-Hydrodynamic seal flow within supercritical carbon dioxide turbomachinery with good computational efficiency.

In this work, we explore the possibility and potential of employing PINNs in approximating the compressor flowfield. To the authors' best knowledge, this is the first attempt to reconstruct the compressor cascade flowfield using PINNs. To this aim, two PINNs were established to solve the forward and inverse problems for the representative 2D subsonic linear compressor cascade flow. Besides, adaptive weights and dynamically adjusting learning rate strategies were implemented to address the convergence challenges arising from gradient imbalances during the training of PINNs. For the forward problem, the Dirichlet boundary conditions were generated based on the CFD data on the boundaries of the computation domain, and the predicted cascade flowfiled by PINNs were compared to the exact values from RANS results. More challenging work is then conducted to solve the inverse problem using PINNs where the cascade flowfield was inferred based on incomplete information of boundary conditions. The evaluation of PINN capability of solving inverse problems in the presence of aleatory uncertainties stemming from lebeled data was conducted at the end. The paper is organized as follows: Section 2 shows the principles and establishment of PINNs. Section 3 presents the CFD and PINN results in solving the forward and inverse problems for simulating compressor cascade flow. The major conclusions and discussions are summarized at the end of this paper.

**II. Methodology**

Here the governing equations of the incompressible compressor cascade flow in the 2D Cartesian reference frame $(x, y)$ are as follows:

$$\partial_t \begin{pmatrix} 0 \\ u \\ v \end{pmatrix} + \partial_x \begin{pmatrix} u \\ u^2 + p/\rho \\ uv \end{pmatrix} + \partial_y \begin{pmatrix} v \\ uv \\ v^2 + p/\rho \end{pmatrix} + \nu \nabla^2 \begin{pmatrix} 0 \\ u \\ v \end{pmatrix} = \mathbf{0} \qquad (1)$$

where $t$ means the time, $(x, y) \in \Omega \subset \mathbb{R}^2$ represents the space coordinates, $u$ represents the axial component of velocity, $v$ means the vertical component of velocity, $p$ is the pressure, $\nu$ means the kinematic viscosity of air, $\nabla^2$ represents the Laplace operator. A constant air density of $1.225\ kg/m^3$ is set here. Because the main interest of this work is the steady state prediction of cascade flow, the time derivate term is omitted in the following part of this paper.

## A. PINNs for Navier-Stokes Equations

To simplify the equation expression, the Eqn. (1) can be re-expressed as:

$$\boldsymbol{f}(\mathbf{x}, \mathbf{u}, p, \partial_x \mathbf{u}, \partial_x p, \dots, \boldsymbol{\lambda}) = 0, \mathbf{x} \in \Omega$$

$$\mathbf{u}(\mathbf{x}) = \boldsymbol{I}(\mathbf{x}), \mathbf{x} \in \Omega \quad (2)$$

$$\mathbf{u}(\mathbf{x}) = \boldsymbol{B}(\mathbf{x}), \mathbf{x} \in \partial\Omega$$

where $\boldsymbol{f}$ donates the residual of the N-S equations containing the differential operators of the velocity $\mathbf{u}$ and pressure $p$ in conjunction with the other parameters $\boldsymbol{\lambda}$ that might be inferred in the inverse problem, $\mathbf{x}$ means the 2D coordinates inside the computational domain, $\boldsymbol{I}$ represents the initial condition, and $\boldsymbol{B}$ means the boundary conditions.

A fully connected NN is then established and trained to make the NN outputs $\mathbf{u}_{NN}(\mathbf{x};\boldsymbol{\theta})$ and $p_{NN}(\mathbf{x};\boldsymbol{\theta})$ close to the exact values of $\mathbf{u}(\mathbf{x})$ and $p(\mathbf{x})$. Here $\boldsymbol{\theta}$ means the trainable parameters of PINNs. One natural advantage of NN in embedding the physical laws of which the governing equations can be expressed in the form of PDEs is to utilize the automatic differentiation (AD) approach [21]. Solving Eqn. (2) is then converted into an optimization problem by minimizing the overall loss function $\mathcal{L}$ on the pre-sampled training points:

$$\mathcal{L}(\boldsymbol{\theta}) = \omega_1 \mathcal{L}_e(\boldsymbol{\theta}) + \omega_2 \mathcal{L}_f(\boldsymbol{\theta}) + \omega_3 \mathcal{L}_b(\boldsymbol{\theta}) \quad (3)$$

where $\mathcal{L}_e$ represents the residual of Eqn. (1), $\mathcal{L}_f$ is the prediction loss on the training points within reference flowfield and $\mathcal{L}_b$ means the training loss on computation boundary, $\omega_i (i = 1, 2, 3)$ are the weights of each loss term. The residual loss of governing equations can be evaluated as follows:

$$\mathcal{L}_e(\boldsymbol{\theta}) = MSE(\boldsymbol{f}, \mathbf{0}) = \frac{1}{N_e}\sum_{\mathbf{x}\in\Omega_e}\|\boldsymbol{f}(\mathbf{x}, \mathbf{u}_{NN}, p_{NN}, \partial_x \mathbf{u}_{NN}, \partial_x p_{NN}, \dots, \boldsymbol{\lambda})\|_2^2 \quad (4)$$

where $MSE$ means the mean square error of governing equation residuals, $N_e$ represents the points number of sample points inside the computation domain, $\Omega_e$ means the dataset of the residual points. The prediction loss $\mathcal{L}_f$ on training points is formulated as

$$\mathcal{L}_f(\boldsymbol{\theta}) = MSE(\mathbf{u}_{NN}(\mathbf{x};\boldsymbol{\theta}), \mathbf{u}(\mathbf{x})) + MSE(p_{NN}(\mathbf{x};\boldsymbol{\theta}), p(\mathbf{x})), \mathbf{x} \in \Omega_e \quad (5)$$

The loss function on the boundary conditions equals the sum of the losses on inlet surface $\mathcal{L}_{b,in}$, losses on outlet surface $\mathcal{L}_{b,out}$, losses on wall surface $\mathcal{L}_{b,wall}$ and losses on periodic surfaces $\mathcal{L}_{b,per}$. The inlet loss is defined as the $L^2$ error of difference between the PINNs-predicted inlet velocity components and the imposed inlet velocity in CFD setups, which yields in form of

$$\mathcal{L}_{b,in} = MSE\big(\mathbf{u}_{NN}(\mathbf{x};\boldsymbol{\theta}), \mathbf{u}(\mathbf{x})\big), \mathbf{x} \in \Omega_{b,in} \quad (6)$$

where $\Omega_{b,in}$ donates the dataset of point coordinates on inlet boundary. $\mathcal{L}_{b,out}$ can be obtained by calculating the $L^2$ error between the PINNs-predicted output pressure and the imposed outlet pressure as follows:

$$\mathcal{L}_{b,out} = MSE\big(p(\mathbf{x};\boldsymbol{\theta}), p(\mathbf{x})\big), \mathbf{x} \in \Omega_{b,out} \quad (7)$$

where $\Omega_{b,out}$ means the dataset of point coordinates on outlet surface. On the wall surface, the non-slip assumption is added, which means the flow velocity on wall surface is fixed to be zero with a loss function in form of:

$$\mathcal{L}_{b,wall} = MSE(\mathbf{u}_{NN}(\mathbf{x};\boldsymbol{\theta}), 0), \mathbf{x} \in \Omega_{b,wall} \quad (8)$$

where $\Omega_{b,wall}$ represents the coordinate dataset on the wall surface. The flow parameters on the upper periodic surface keep the same as the lower surface with the definition of loss:

$$\mathcal{L}_{b,per} = MSE\big(\mathbf{u}_{NN}(\mathbf{x}_{lower};\boldsymbol{\theta}), \mathbf{u}_{NN}(\mathbf{x}_{upper};\boldsymbol{\theta})\big), \mathbf{x}_{upper} \cup \mathbf{x}_{lower} \in \Omega_{b,per} \quad (9)$$

where $\Omega_{b,per}$ means the coordinate dataset on periodic boundary. The structure of the established PINNs is shown in Fig. 1. The RANS governing equations are deployed instead of Eqn. (1) because the training data required by PINNs come from the RANS results in this paper. In detail, the mean flow parameters $\bar{u}, \bar{v}, \bar{p}$ were solved and the new viscosity is calculated by adding up the laminar $\nu$ and turbulent viscosity $\nu_{tur}$. The specific values of the turbulent viscosity are calculated through the deployment of turbulent models (detailed in the CFD setups). In this study, the Tanh activation function is employed, and the neurons are fully connected across the layers. The aforementioned loss equations are incorporated into the overall loss function of the PINNs.

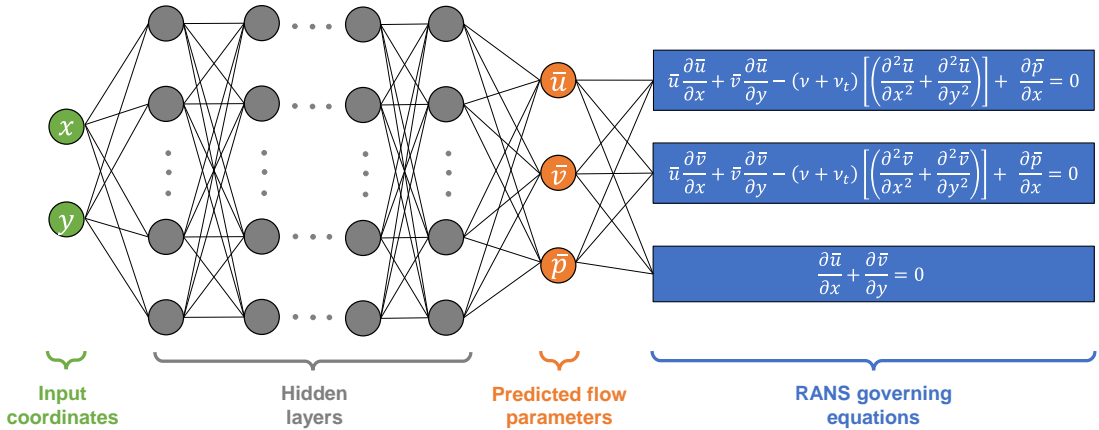

**Fig. 1 Structure of PINNs.**

**B. CFD Setups**

The NACA-65 profile-based linear compressor cascade, which is supported by the well-documented experimental measurements conducted by Ma [22] at École Centrale de Lyon, is selected as the reference geometry. The key design parameters are summarized in Tab. 1.

**Table 1 Key parameters of compressor cascade**

| Parameter | Value |
| --- | --- |
| Solidity | 1.12 |
| Aspect ratio | 2.47 |
| Chord length | 0.15 m |
| Span length | 0.37 m |
| Inlet velocity magnitude | 40.00 m/s |
| Stagger angle | 42.70° |
| Camber angle | 23.22° |
| Design upstream flow angle | 54.31° |
| Design downstream flow angle | 31.09° |

The configuration of the computational domain and the boundary conditions is depicted in Figure 2a. The domain is extended 2.9 times and 3.7 times the axial chord length upstream and downstream of the compressor cascade, respectively. On the inlet plane, a velocity magnitude of 40 m/s with an incidence angle of 4° is imposed. The outlet gauge pressure is set to 0 Pa. To create a periodic effect, the translational periodic boundary condition is applied to the lateral surfaces of the computational domain, with a tangential pitch of 0.134 m. The compressor cascade surface is treated with a non-slip wall condition. The mesh for the passage within the compressor cascade is generated using ANSYS/ICEM-CFD. The H-O-H grid topology is utilized to discretize the spatial domain of the cascade passage, while an O-topology is generated near the cascade wall. The height of the first cell near the wall surface is set to 1e-5 m to ensure that the maximum value of $y^+$ remains below 1 across the entire computational domain. Figure 2b illustrates a logarithmic plot depicting the convergence of prediction errors in relation to changes in mesh size. The horizontal axis represents the average mesh size $h$, while the vertical axis represents the disparity between the calculated pressure loading and the experimental measurement data. The chosen total element count in the two dimensions is 130,000. The mesh visualization produced is depicted in Figure 2c.

The 2D RANS equations are solved to simulate the compressor cascade flow using open-source CFD code SU2 [23]. SU2 is a computational analysis and design package that aims to solve multiphysics analysis and optimization problems. Its primary applications are CFD and aerodynamic shape optimizations with the support of continuous and discrete adjoint calculations. The pressure-velocity coupling scheme is deployed here to update the flow pressure and velocity simultaneously in each calculation step. For spatial discretization, the finite volume method with a second-order upwind scheme is used for solving the convection terms in the governing RANS equations. The one-equation Spalart-Allmaras (SA-noft2) [24] is deployed here to solve the transport equation of kinematic eddy turbulent viscosity. The CFD calculations are regarded as convergent when the maximal residual value becomes smaller than 1e-6.

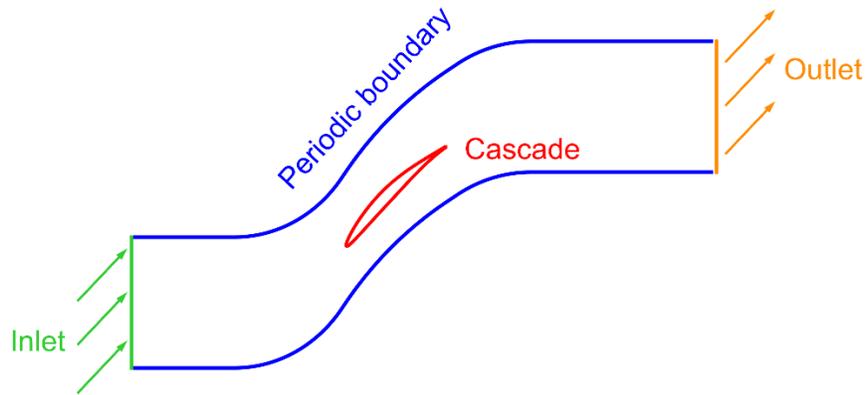

a)

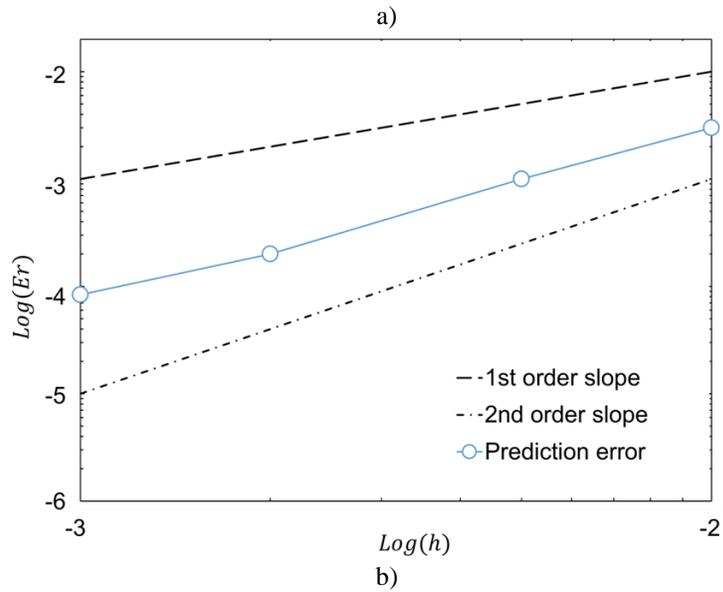

b)

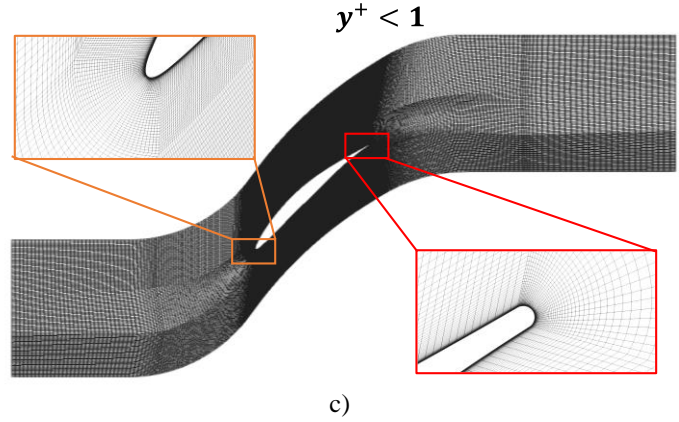

c)

**Fig. 2 Computation domain and mesh: a) Boundary conditions; b) Log-log plot of error convergence; c) Generated mesh.**

## III. CFD and PINN Results

The numerical simulation results are first compared to the available experimental data to guarantee the CFD data accuracy before transmitting to the training dataset for PINNs. The following section focuses on the simulation case with an incidence angle of 4°.

### A. CFD Results

The comparison of near-wall static pressure coefficients $C_p$ are shown in Fig. 3. The solid line represents the CFD results, and the scatter points means the experiment data. The definition of $C_p$ is:

$$C_p = \frac{p_{s\_nw} - p_{s\_in}}{P_{t\_in} - p_{s\_in}} \tag{10}$$

where $p_{s\_nw}$ represents the near-wall static pressure at the measurement points, $p_{s\_in}$ the inlet static pressure and $P_{t\_in}$ represents the inlet total pressure. The comparison plot validates that the adopted CFD approach can predict the compressor cascade flow with enough accuracy. In detail, the streamwise airflow near the cascade suction surface is gradually decelerated, and the air pressure is substantially increased along the streamwise direction.

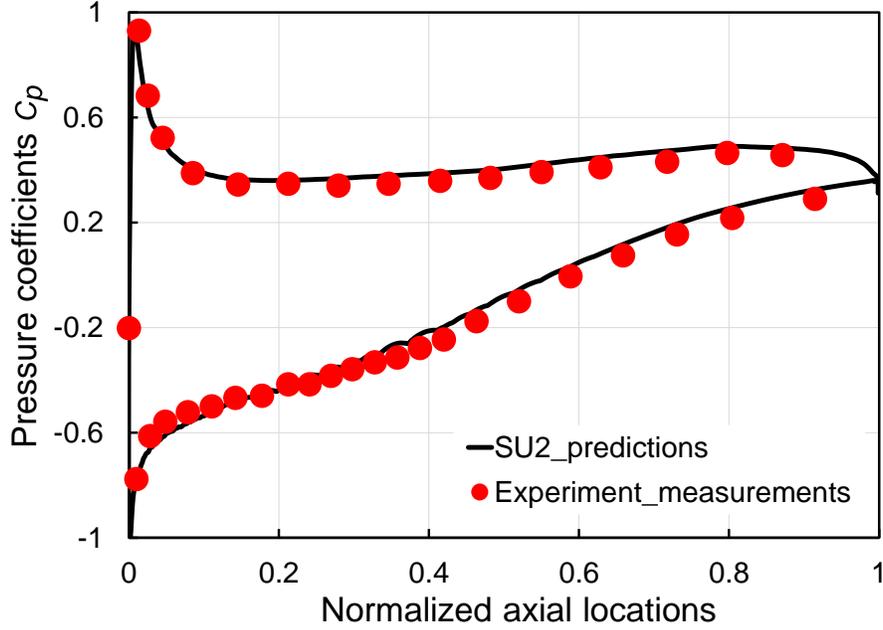

**Fig. 3 Comparison of near-wall static pressure coefficients $C_p$ between CFD and experiment measurement data.**

**B. PINN Results-Forward Problem**

To solve the forward problem, 20,000 randomly distributed mesh points within the computation domain are specified as residual points. In addition, the CFD solution points on the inlet (108 points), outlet (108 points) and wall (553 points) surfaces are specified as Dirichlet boundary conditions. On the periodic surfaces, the flow parameters on 814 upper- and lower-point pairs are constrained to be the same. 2,000 randomly distributed mesh points are selected as labeled data points. The distributions of these points are shown in Fig. 4. Here the residual and labeled points are non-uniformly distributed and clustered near the cascade wall, which is due to the denser mesh generated near the wall. Having used uniform sampling based on grid nodes, those sampled points cluster near the wall of the cascade (where the grid is denser). The fully connected NNs consist of 9 hidden layers with [20, 50, 100, 100, 200, 200, 100, 50, 20] neurons on each layer. The input layer consists of 2 neurons, one for the axial coordinate dataset $x$ and one for the vertical coordinate dataset $y$. The output layer contains 3 neurons, one for the axial component of velocity $u$, one for the vertical component of velocity $v$ and one for static pressure $p$. An alternative output layer is to use a stream function instead of directly outputting the velocity components, which automatically satisfies the mass conservation requirement. However, our test results show that the usage of stream function does not lead to an improvement in the PINNs prediction, but rather to some non-physical phenomenon near the leading edge. The potential cause might be the incorporation of third-order gradient computations into the back-propagation phase during training, which negatively impacts the convergence of PINNs. Thus, the output layer with the velocity components is deployed in the following work.

A significant constraint faced by vanilla PINNs lies in their limited assurance of convergence due to the imbalanced gradients during the back-propagation phase between the PDE residual term and the labeled training loss term in the model training procedure, as highlighted in references [25, 26]. To enhance the convergence effectiveness in the optimization of PINNs, adaptive weights for the loss terms are introduced to naturally equilibrate the gradients during back-propagation. Differing from the approach in reference [25] where the gradient elements of the PDE residual loss are chosen using a maximization operation, this study employs the mean operation to yield improved convergence performance. The algorithm of generating adaptive weights is as following:

$$\mathcal{L}(\boldsymbol{\theta}) = \omega_1 \mathcal{L}_e(\boldsymbol{\theta}) + \omega_2 \mathcal{L}_f(\boldsymbol{\theta}) + \omega_3 \mathcal{L}_b(\boldsymbol{\theta}),$$

$$\omega_1 = 1,$$

$$\omega_2 = \text{mean}_{\boldsymbol{\theta}} \left\{ \frac{|\nabla_{\boldsymbol{\theta}} \mathcal{L}_e(\boldsymbol{\theta})|}{|\nabla_{\boldsymbol{\theta}} \mathcal{L}_f(\boldsymbol{\theta})|} \right\}, \quad (11)$$

$$\omega_3 = \text{mean}_{\boldsymbol{\theta}} \left\{ \frac{|\nabla_{\boldsymbol{\theta}} \mathcal{L}_e(\boldsymbol{\theta})|}{|\nabla_{\boldsymbol{\theta}} \mathcal{L}_b(\boldsymbol{\theta})|} \right\},$$

where $\nabla_{\boldsymbol{\theta}}$ donates the gradient matrix for specific loss term to the model parameters.

The weights are initialized to be normally distributed using Glorot initialization algorithm [27], and the values of biases are initialized to be zero. Besides, to accelerate the convergence of PINNs, the chained scheduler of dynamically adjusting learning rate is adopted in the training loops. In detail, the warmup scheme is first deployed in the first 10 steps where the learning rate starts with a very small value ($lr_{min}$ =1e-07) and is then gradually increased to the maximum value ($lr_{max}$ =1e-03). The cosine decaying scheme is then applied by periodically decreasing the learning rate in the form of

$$lr = lr_{min} + 0.5(lr_{max} - lr_{min})\left(1 + \cos\left(\pi(T_{cur}^i/T^i)\right)\right), T^i = 100 * 2^i, i = 1,2,..,n \quad (12)$$

where $T_{cur}^i$ means the current epoch steps within the $i$-th running cycle of the decaying period, $T^i$ is the total epoch number of the $i$-th cycle of the decaying period, $n$ donates the total number of decaying cycles. Researchers interested in abovementioned learning rate adjustment algorithm can refer to Ref. [28] for more details.

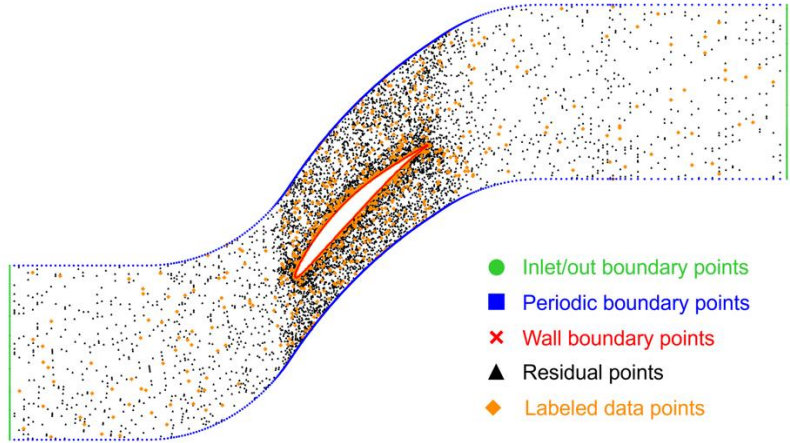

**Fig. 4 Training dataset of PINNs for forward problem.**

The *PyTorch* [29] environment is used to generate the PINNs and the training process was carried out on one graphics processing unit (GPU) Nvidia RTX 6000 card of the High-Performance Computing (HPC) cluster at Imperial College London. The history of training loss is illustrated in Fig. 5. The overall training error shows a cyclical downward trend due to the warmup-cosine-decay mechanism in adjusting the learning rate, and the total loss reaches $\mathcal{O}(10^{-4})$. Among these losses, the residual loss accounts for the largest proportion, followed by the losses on the labeled data.

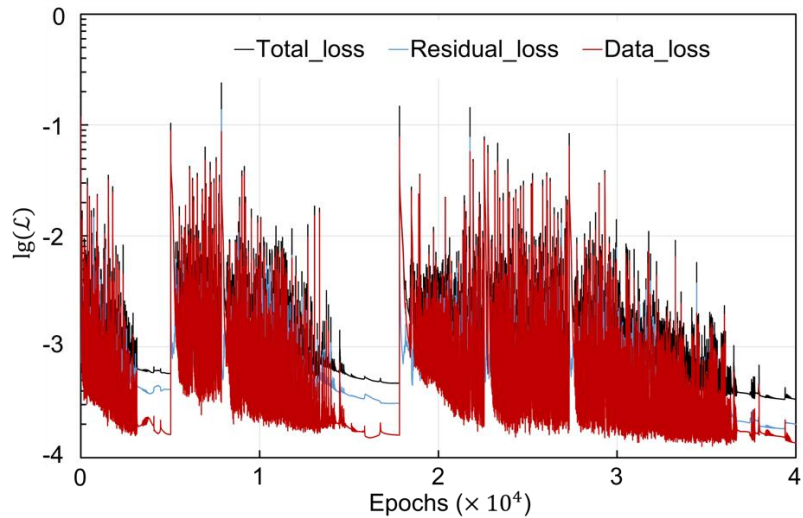

**Fig. 5 Training history of PINNs for forward problem.**

The contour plots of flow velocity and pressure were generated using the well-trained PINNs and then compared to the CFD data. The comparison results are summarised in Fig. 6. In terms of the pressure field, the PINNs exhibit strong predictive capability, accurately representing the static pressure distribution within the compressor cascade passage. Furthermore, the PINN predictions for the distributions of axial and vertical velocity components closely align with the CFD results. Notably, the wakes generated by the adjacent cascade are also reproduced by the PINNs, as observed in the figure. Fig. 7 focuses on the comparison of near-wall static pressure coefficients between the CFD results and PINN predictions. The pressure difference between the suction surface and pressure surface of compressor cascade is well captured by PINNs, and thus the airflow is then "compressed" through the compressor. This comparison highlights the PINNs' ability to accurately approximate flow details along the cascade wall surface. Tab. 2 provides a summary of both the absolute and relative Mean Square Error (MSE) values for the various flow parameters predicted by the PINN model. The maximum absolute MSE value is associated with the vertical velocity field prediction, and this might be because the airflow has a larger component in the vertical direction under the condition of a positive incidence angle. The maximum relative MSE value reaches 2.39% for the prediction of vertical velocity. The relative prediction error for static pressure and axial vertical velocity is around 0.22% and 0.79%, respectively. All $R^2$ values are larger than 0.99, indicating that the developed PINN model effectively explains the variation in the flow parameters.

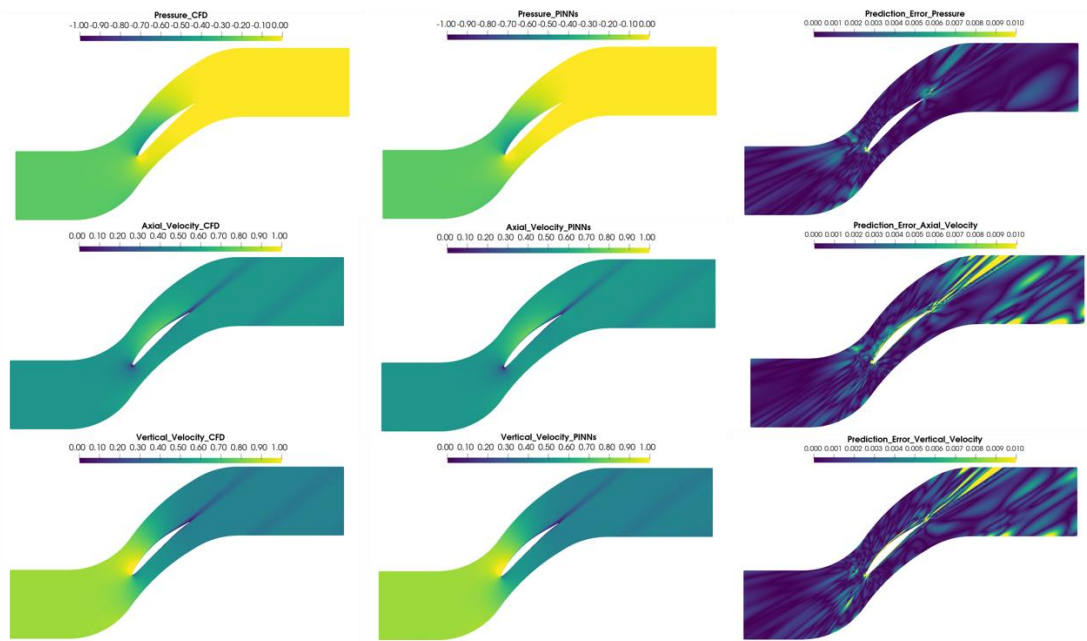

**Fig. 6 Contour plot comparison between CFD and PINNs results for forward problem.**

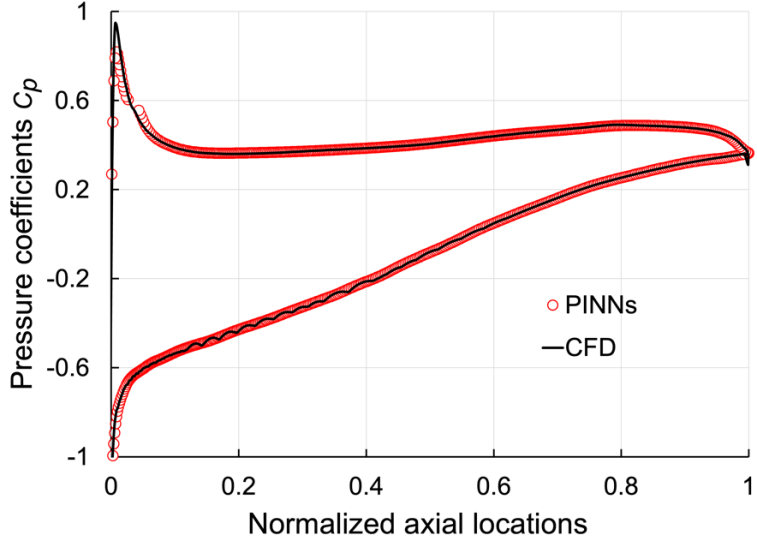

Fig. 7 Comparison of near-wall static pressure coefficients $C_p$ between CFD and PINNs results

Table 2 Predictive error of PINNs for forward problem

|  | Pressure | Axial Velocity | Vertical Velocity |
| --- | --- | --- | --- |
| Absolute MSE for PINNs | 5.94e-05 | 3.58e-04 | 1.83e-03 |
| Relative MSE for PINNs | 0.22% | 0.79% | 2.39% |
| R-squared value ($R^2$) | 0.99 | 0.99 | 0.99 |

Besides, the conventional deep neural networks (DNNs) featured by the same architecture as PINNs are then built to compare their capabilities in predicting the compressor flowfield, as shown in Fig. 8. The only difference between these two networks is that the governing equation loss is not included in the total loss definition for the DNNs. The Tanh activation function is employed here, and the neurons are fully connected across the layers.

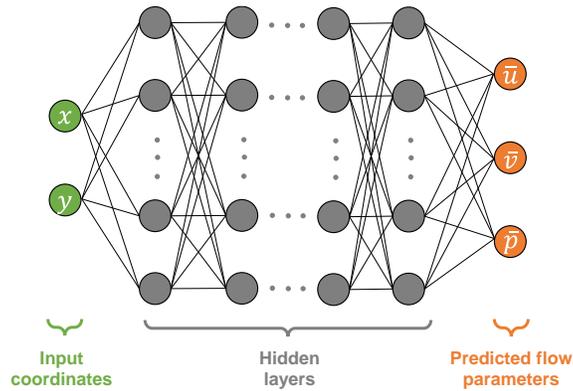

Fig. 8 Structure of DNNs.

The comparison results between the PINNs and DNNs are shown in Fig. 9. In detail, the DNNs produce a flow field that exhibits unrealistic stripe patterns. This phenomenon could be attributed to the absence of a locality constraint. Incorporating convolutional operations, as commonly employed in Convolutional Neural Networks (CNNs), may prove beneficial in enhancing the performance of the neural networks without imposing physical constraints. In comparison, the flow solutions predicted by PINNs are much smoother. The reason is that the PINNs can interpolate the flow parameters on the mesh points where the solutions are unknown with the aid of the hidden information within the governing equations. The quantitative comparison between the PINNs and DNNs is conducted in terms of the absolute and relative MSE, and the comparison results are summarised in Tab. 3. Apparently, PINNs outperform DNNs in terms of prediction accuracy of compressor cascade flow for the same training dataset. The relative MSE of vertical velocity that predicted by PINNs is reduced by a maximum of 0.79% when compared to the DNNs.

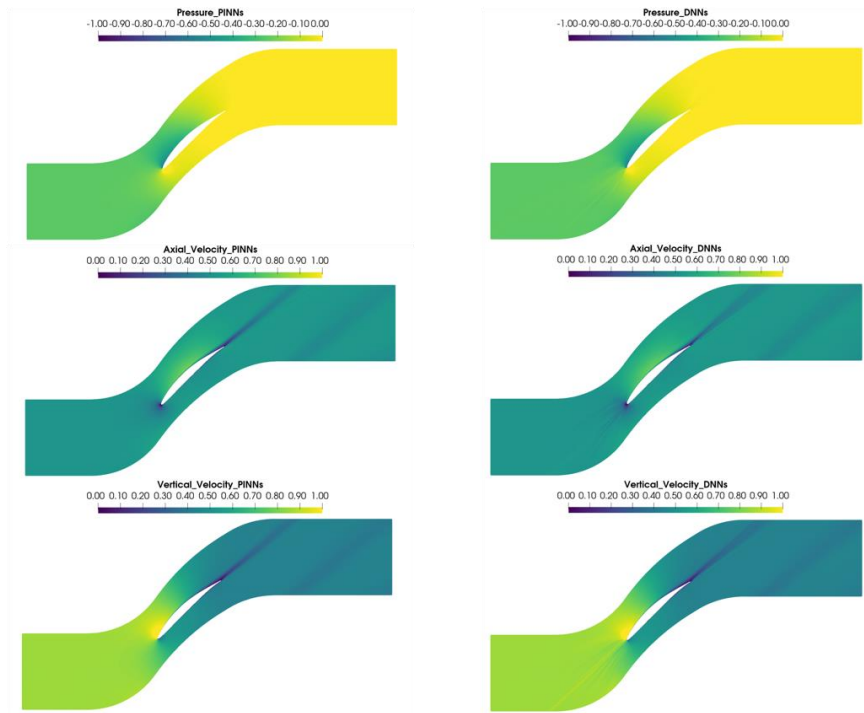

**Fig. 9 Comparison of contour plots between PINNs and DNNs for forward problem.**

**Table 3 Mean square error of DNNs for forward problem**

|  | Pressure | Axial Velocity | Vertical Velocity |
|---|---|---|---|
| Absolute MSE for DNNs | 5.98e-05 | 6.68e-04 | 2.44e-03 |
| Absolute MSE for {DNNs}-Absolute MSE for {PINNs} | +4.00e-07 | +3.10e-04 | +6.10e-04 |
| Relative MSE for DNNs | 0.23% | 1.48% | 3.18% |
| Relative MSE for {DNNs}-Relative MSE for {PINNs} | +0.01% | +0.69% | +0.79% |

### C. PINN Results-Inverse Problem

The inverse problem refers to the calculation of a flow field based on a subset of what is defined at the observation database. In many engineering applications, it's common for the boundary conditions to be either unknown or challenging to determine precisely. One potential application of PINNs for solving inverse problems about turbomachinery is the reconstruction of inner flow field while some discrete/sparse boundary data may be available. Assuming that the local velocity vectors can be extracted using particle imaging velocimetry (PIV), and meanwhile, the static pressure on the cascade wall surface can be measured using pressure probes, it would be promising to reconstruct the complete cascade flowfield using PINNs. To achieve this goal, 20,000 randomly distributed mesh points within the computation domain are specified as residual points. In addition, 553 mesh points (only static pressure information available) on the cascade wall surface and 1,000 randomly distributed mesh points (only velocity components information available) within the computation domain are selected as the labeled data points. The other inlet/outlet/wall/periodic boundary conditions are set to unknown. The distributions of the residual and labeled points are shown in Fig. 10.

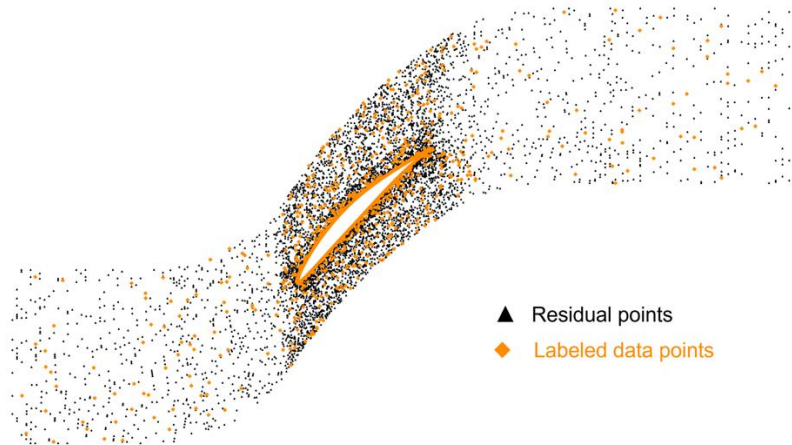

**Fig. 10 Training dataset of PINNs for inverse problem.**

The same NN architecture is used here for solving the forward problem. The adaptive weight strategy is adopted and the learning rate is dynamically adjusted through the warmup-cosine-decay scheme. The PINNs are established in the environment of *PyTorch* and the whole training process is conducted using GPU Nvidia RTX 6000 in HPC at Imperial. The history of training loss is plotted in Fig. 11. Even though there is much less information compared to the forward problem, the overall training loss of PINNs reaches $\mathcal{O}(10^{-3})$. Within this loss breakdown, the predominant component is the residual loss stemming from the governing equations, constituting the most substantial proportion. Subsequently, the losses associated with labeled data are of a lower magnitude, approximately at $\mathcal{O}(10^{-4})$.

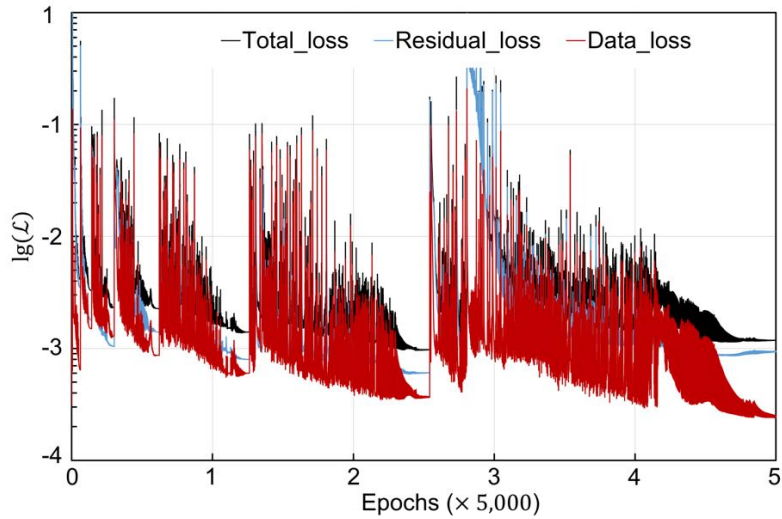

Fig. 11 Training history of PINNs for inverse problem.

Figure 12 presents a comparison of contour plots between the PINNs and CFD results. Remarkably, PINNs demonstrate the ability to accurately predict the pressure field even in the absence of labeled pressure data within the compressor cascade passage. This proficiency in static pressure prediction arises from the PINNs' capability to infer governing equations. Moreover, the velocity field predicted by PINNs exhibits a flow pattern akin to that observed in the CFD results. PINNs prove adept at faithfully capturing the wake generated by the adjacent cascade. The distributions of near-wall static pressure coefficients in Fig. 13 further validates the effectiveness of PINNs in approximating the flow characteristics along the cascade wall surface. The various MSE errors on the sample data are collected in Tab. 4 where the maximum absolute prediction occurs in predicting the vertical velocity field, with the maximum relative MSE error of 4.06%.

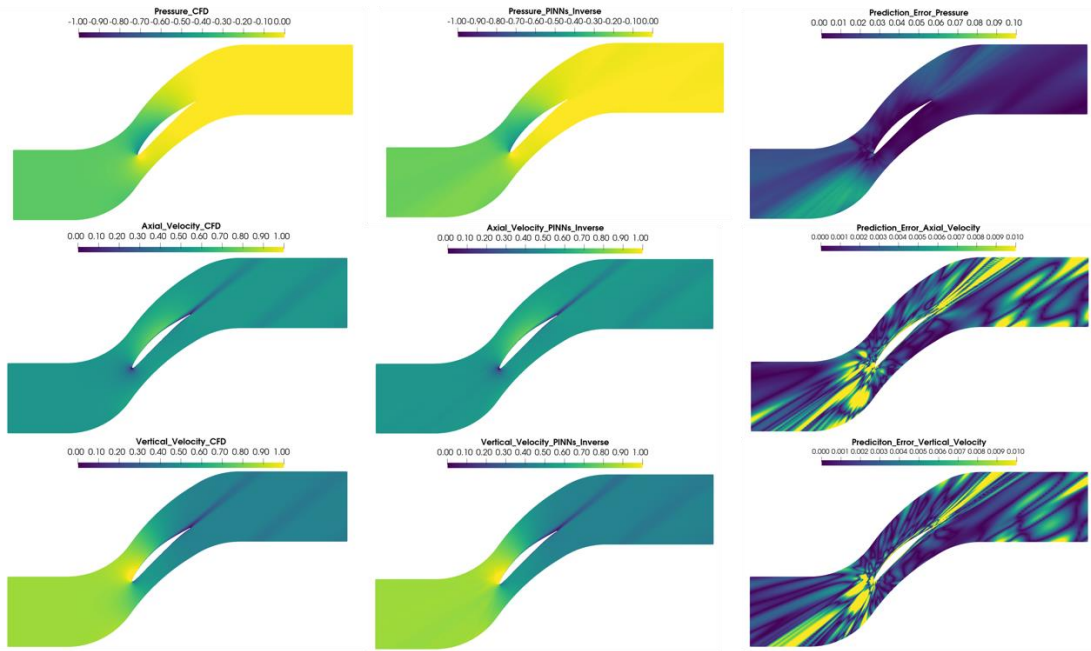

**Fig. 12 Contour plot comparison between CFD and PINNs results for inverse problem.**

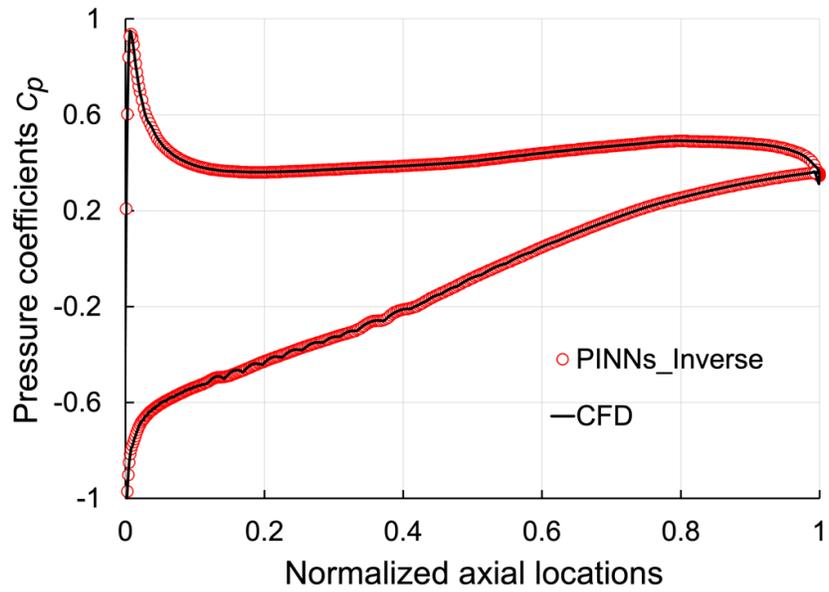

**Fig. 13 Comparison of near-wall static pressure coefficients $C_p$ between CFD and inverse PINNs results.**

Table 4 Mean square error of PINNs for inverse problem

|  | Pressure | Axial Velocity | Vertical Velocity |
|---|---|---|---|
| Absolute MSE for PINNs | 2.06e-04 | 5.03e-04 | 3.12e-03 |
| Relative MSE for PINNs | 0.75% | 1.11% | 4.06% |
| R-squared value ($R^2$) | 0.99 | 0.99 | 0.99 |

The PINNs' performance is then compared to the DNNs, and the comparison results is shown in Fig. 14. DNNs lose their capability in predicting pressure field in the absence of prior training dataset of pressure. Similar to the situation in solving forward problem, PINNs show a better performance in interpolating the flowfield where the solutions of mesh points are not known when compared to the DNNs. The MSE comparison collected in Tab. 5 validates that the prediction performance of NNs in the inference of unlabeled information can be substantially enhanced with the physical constraints imposed.

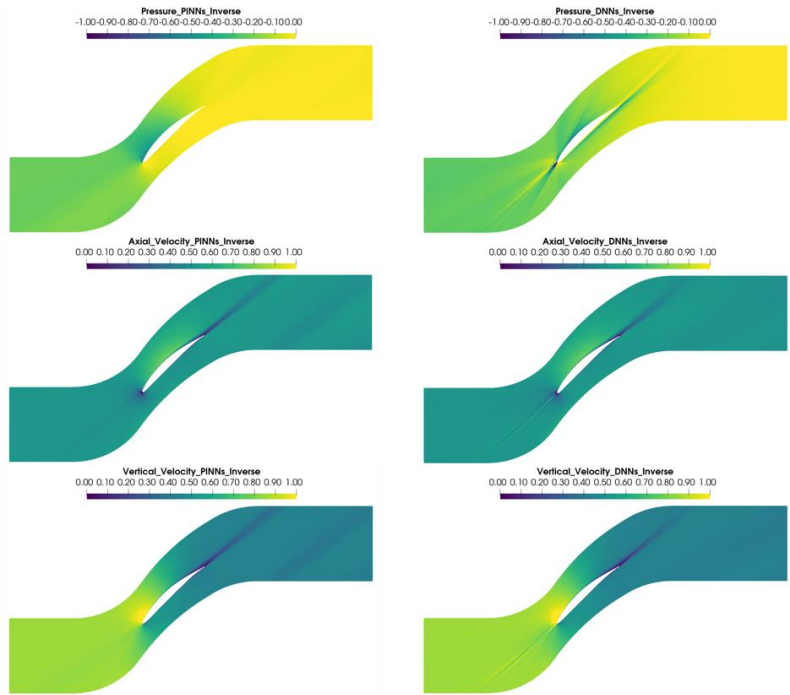

Fig. 14 Comparison of contour plots between PINNs and DNNs for inverse problem.

Table 5 Mean square error of DNNs for inverse problem

|  | Pressure | Axial Velocity | Vertical Velocity |
|---|---|---|---|
| Absolute MSE for DNNs | 1.31e-02 | 7.44e-04 | 3.13e-03 |
| Absolute MSE for {DNNs}- Absolute MSE for {PINNs} | +1.29e-02 | +2.41e-04 | +1.00e-05 |
| Relative MSE for DNNs | 47.64% | 1.65% | 4.08% |
| Relative MSE for {DNNs}-Relative MSE for {PINNs} | +46.89% | +0.54% | +0.02% |

To assess the impact of aleatory uncertainties, various levels of noise are introduced to model data variability stemming from experimental measurements or numerical simulation errors. Three distinct levels of uncertainty, assumed to follow Gaussian distributions with a mean of zero, are incorporated. These uncertainties span a range of 1%, 5%, and 10% relative to the reference velocity and pressure values. The architecture and training process of the PINNs remain consistent with the descriptions provided earlier. The relative MSE for both velocity vectors and pressure serves as the performance indicator for the PINNs. Fig. 15 illustrates how the PINNs respond to the introduced uncertainties. As the level of aleatory uncertainty increases, there is a gradual enhancement in the relative MSE for velocity prediction. The relative MSE values of pressure fluctuate within an even narrower range when compared to the velocity prediction. One interesting phenomenon is that, compared to the baseline PINNs, the prediction error is diminished after the uncertainty magnitude is increased to 10%. The reduction in MSE values could be attributed to a potential reduction in overfitting after introducing uncertainty, such as dropout mechanisms in Bayesian NNs. Notably, the variation range of MSE values remains below 1.2% for all uncertainty levels, demonstrating the robustness of the PINNs in solving inverse problems while accounting for varying levels of aleatory uncertainties. This is further confirmed by comparing the flowfields, as illustrated in Fig. 16, where the PINNs incorporating a 10% uncertainty on labeled data predict nearly identical flow patterns to the baseline PINNs.

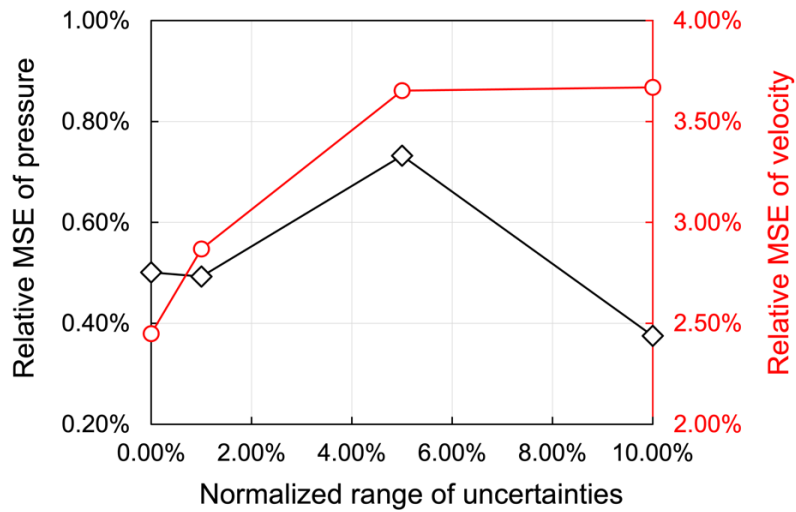

**Fig. 15 Prediction of PINNs for inverse problems with various levels of uncertainties from labeled data.**

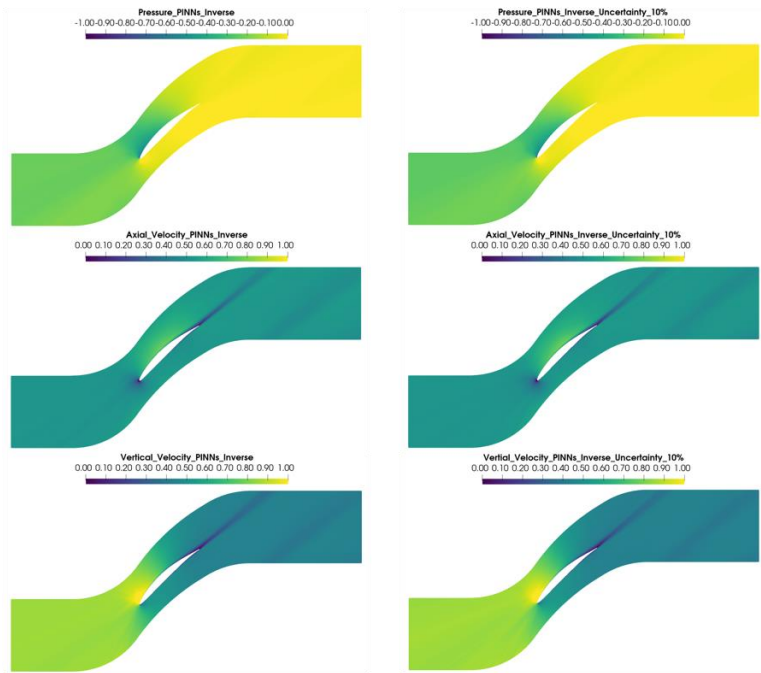

**Fig. 16 Comparison of contour plots between PINNs with/without uncertainty for inverse problem.**

Velocity components downstream of the reference cascade play a significant role in influencing the performance of downstream stages within multistage turbomachinery. Consequently, the velocity components on the transverse line corresponding to 0.1 times the chord length downstream cascade are extracted and analysed here. The detailed comparisons of velocity magnitude and angle between the results obtained from CFD and PINNs with and without external uncertainties are illustrated in Fig. 17. Regarding velocity magnitude, the predictions from both sets of PINNs closely align with the CFD results. The velocity depict attributed to the blade body is well captured by PINNs. Concerning flow angle distributions, while there is a slight increase in prediction error when considering PINNs with uncertainty compared to the baseline PINNs, the model still effectively replicates the distribution patterns downstream of the cascade trailing edge. The maximum prediction error of velocity angle remains below 1.4%. These results provide further evidence of the effectiveness of PINNs in the turbomachinery design process.

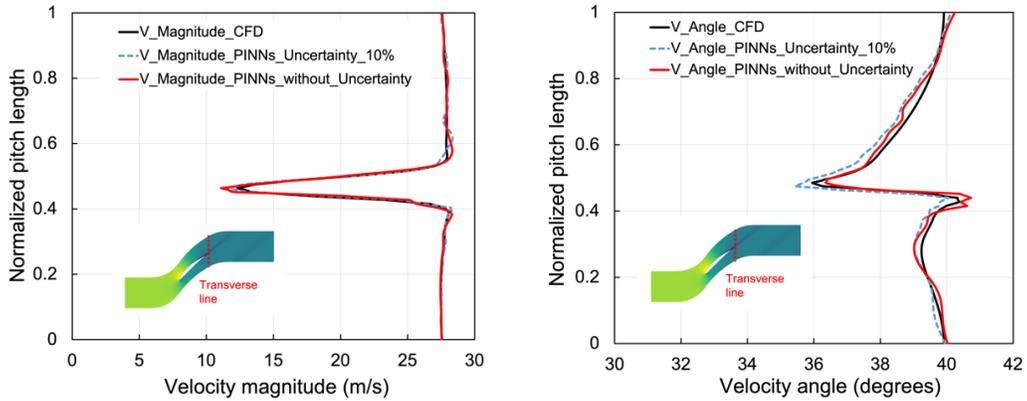

**Fig. 17 Comparison of velocity components between PINNs with/without uncertainty downstream cascade.**

## IV. CONCLUSIONS

The current research work utilizes physics-informed neural networks (PINNs) to simulate the flowfield within a compressor cascade passage. The following key conclusions can be drawn from the study:

1) PINNs effectively solve the governing Navier-Stokes (N-S) equations for the forward problem by leveraging labeled training points within the computation domain and on the boundary. The flowfield predicted by PINNs demonstrates good agreement with CFD results.

2) PINNs exhibit advantages over traditional CFD approaches when addressing the inverse problem, especially in the absence of boundary conditions. By utilizing local velocity vectors and near-wall pressure information, PINNs accurately reconstruct the global flowfield.

3) Compared to deep neural networks (DNNs), PINNs demonstrate enhanced capabilities in obtaining reasonable compressor cascade flowfields for both positive and inverse problems. Besides, PINNs demonstrate robustness in solving inverse problems while accounting for varying levels of aleatory uncertainties.

## Funding Sources

The authors would like to sincerely thank the support of the grant from the European Union's Marie Skłodowska-Curie Actions Individual Fellowship (MSCA-IF-MENTOR-101029472).